\documentclass{elsarticle}
\usepackage[english]{babel}
\usepackage{amssymb}
\usepackage{makeidx}  
\usepackage{rotating}
\usepackage{arydshln}
\usepackage{amsmath}
\usepackage{hyperref}
\usepackage{multirow}

\usepackage[monochrome]{color}

\journal{Computer Vision and Image Understanding}

\begin{document}

\begin{frontmatter}

\title{SR-Clustering: Semantic Regularized Clustering for Egocentric Photo Streams Segmentation}
 
\author[UB,CVC]{Mariella Dimiccoli\corref{cor1}}
\ead{mariella.dimiccoli@cvc.uab.es}
\author[UB]{Marc Bola{\~n}os\corref{cor1}}
\ead{marc.bolanos@ub.edu}
\author[UB,RUG]{Estefania Talavera}
\ead{etalavera@ub.edu}
\author[UB]{Maedeh Aghaei}
\ead{maghaeigavari@ub.edu}
\author[SOF]{Stavri G. Nikolov}
\ead{stavri.nikolov@imagga.com}
\author[UB,CVC]{Petia Radeva\corref{cor2}}
\ead{petia.ivanova@ub.edu}
\cortext[cor1]{Corresponding authors. The first two authors contributed equally to this work.}
\cortext[cor2]{Principal corresponding author}
\address[UB]{Universitat de Barcelona, Barcelona, Spain}
\address[RUG]{University of Groningen, Groningen, Netherlands}
\address[CVC]{Computer Vision Center, Barcelona, Bellaterra, Spain}
\address[SOF]{Imagga Technologies Ltd and Digital Spaces Living Lab, Sofia, Bulgaria}
\begin{abstract}
While wearable cameras are becoming increasingly popular, locating relevant information in large unstructured collections of egocentric images is still a tedious and time consuming process. This paper addresses the problem of organizing egocentric photo streams acquired by a wearable camera into semantically meaningful segments, hence  making an important step towards the goal of automatically annotating these photos for browsing and retrieval.
In the proposed method, first, contextual and semantic information is extracted for each image by employing a Convolutional Neural Networks approach. Later, a vocabulary of concepts is defined in a semantic space by relying on linguistic information. Finally, by exploiting the temporal coherence of concepts in photo streams, images which share contextual and semantic attributes are grouped together. The resulting temporal segmentation is particularly suited for further analysis, ranging from  event recognition to semantic indexing and summarization. Experimental results over egocentric set of nearly 31,000 images, show the prominence of the proposed approach over state-of-the-art methods. 
\end{abstract}


\end{frontmatter}


\section{Introduction}

\vspace*{-0.25cm}
\begin{figure}[ht!]
\centering
\includegraphics[width=35em,height=10em]{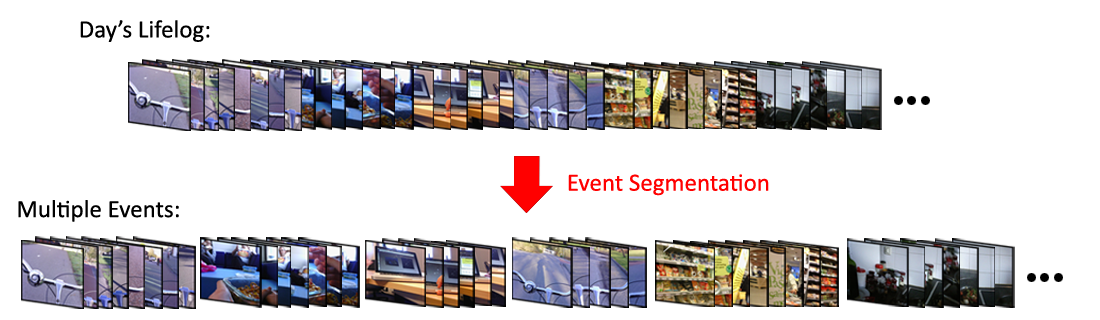}
\caption{Example of temporal segmentation of an egocentric sequence based on what the camera wearer sees. In addition to the segmentation, our method provides a set of semantic attributes that characterize each segment.}
\label{fig:segmentation}
\end{figure}

Among the advances in wearable technology during the last few years, wearable cameras specifically have gained more popularity \cite{bolanos2015towards}. These small light-weight devices allow to capture high quality images in a hands free fashion from the first-person point of view. Wearable video cameras such as GoPro and Looxcie, by having a relatively high frame rate ranging from 25 to 60 fps, are mostly used for recording the user activities for a few hours. Instead, wearable photo cameras, such as the Narrative Clip and SenseCam, capture only 2 or 3 fpm and are therefore mostly used for image acquisition during longer periods of time (e.g. a whole day).
The images collected by continuously recording the user's life, can be  used for understanding the user's lifestyle and hence they are potentially beneficial for prevention of non-communicative diseases associated with unhealthy trends and risky profiles (such as obesity, depression, etc.). In addition, these images can be used  as an important tool for prevention or hindrance of cognitive and functional decline in elderly people \cite{AJM-Wearablecameras}. However, egocentric photo streams generally appear in the form of long unstructured sequences of images, often with high degree of redundancy and abrupt appearance changes even in temporally adjacent frames, that harden the extraction of semantically meaningful content.
Temporal segmentation, the process of organizing unstructured data into homogeneous chapters, provides a large potential for extracting semantic information. Indeed, once the photo stream has been divided into a set of homogeneous and manageable segments, each segment can be represented by a small number of key-frames and indexed by semantic features, providing a basis for  understanding the semantic structure of the event.



\textcolor{red}{State-of-the-art methods for temporal segmentation can be broadly classified into works with focus on what-the-camera-wearer-sees \cite{CastroHickson2015,Doherty:2008,talavera2015r-clustering} and on what-the-camera-wearer-does \cite{polegtemporal,PolegEP015}. As an example, from the what-camera-wearer-does perspective, the camera wearer spending time in a bar while sit, will be considered as a unique event (sitting). From the what-the-camera-wearer-sees perspective, the same situation will be considered as several separated events (waiting for the food, eating, and drinking beer with a friend who joins later). The distinction between the aforementioned points of view is crucial as it leads to different definitions of an event. In this respect, our proposed method fits in the what-the-camera-wearer-sees category.} 
Early works on egocentric temporal segmentation \cite{Doherty:2008,kmean} focused on what the {\em camera wearer sees} (e.g. people, objects, foods, etc.). For this purpose, the authors used as image representation, low-level features to capture the basic characteristics of the environment around the user, such as color, texture or information acquired through different camera sensors.
More recently, the works in \cite{bolanos2015visual} and \cite{talavera2015r-clustering} have used Convolutional Neural Network (CNN) features extracted by using the AlexNet model \cite{NIPS2012_4824} trained on ImageNet as a fixed feature extractor for image representation. 
Some other recent methods infer from the images what the {\em camera wearer does}  (e.g. sitting, walking, running, etc.). Castro et al.  \cite{CastroHickson2015} used CNN features together with metadata and color histogram \cite{CastroHickson2015}.

Most of these methods use as image representation ego-motion \cite{egovideo,bolanos2014video,polegtemporal,PolegEP015}, which is closely \textcolor{red}{related to the user motion-based activity but cannot be reliably estimated in photo streams.}
The authors combined a CNN trained on egocentric data with a posterior Random Decision Forest in a late-fusion ensemble, obtaining promising results for a single user. However, this approach lack of generalization, since it requires to re-train the model for any new user, implying to manually annotate large amount of images. To the best of our knowledge, except the work of Castro et al. \cite{CastroHickson2015}, Doherty et al. \cite{Doherty:2008} and Tavalera et al. \cite{talavera2015r-clustering}, all other state-of-the-art methods have been designed for and tested on videos.
 In our previous work \cite{talavera2015r-clustering}, we proposed an unsupervised method, called \emph{R-Clustering}, \textcolor{red}{aiming to segment photo streams from the what-the-camera-wearer-see perspective}.  The proposed methods relies on the combination of Agglomerative Clustering (AC), that usually has a high recall, but leads to temporal over-segmentation, with a statistically founded  change detector, called ADWIN \cite{bifet2007learning}, which despite its high precision, usually leads to temporal under-segmentation. Both approaches are integrated in a {\em Graph-Cut (GC)} \cite{boykov2001fast} framework to obtain a trade-off between AC and ADWIN, which have complementary properties. The graph-cut relies on CNN-based features extracted using AlexNet, trained on ImageNet, as a fixed feature extractor in order to detect the segment boundaries.

In this paper, we extend our previous work by adding a semantic level to the image representation. 
Due to the free motion of the camera and its low frame rate,  abrupt changes are visible even among temporally adjacent images (see Fig. \ref{fig:segmentation} and Fig. \ref{fig:imagenVariabilidad.jpg}). Under these conditions motion and low-level features such as color or image layout are prone to fail for event representation, hence urges the need to incorporate higher-level semantic information.
Instead of representing images simply by their contextual CNN features, which capture the basic environment appearance, we detect segments as a set of temporally adjacent images with the same contextual representation in terms of semantic visual concepts. Nonetheless, not all the semantic concepts in an image are equally discriminant for environment classification: objects like trees and buildings can be more discriminant than objects like dogs or mobile phones, since the former characterizes a specific environment such as forest or street, whereas the latter can be found in many different environments. In this paper, we propose a method called Semantic Regularized Clustering (SR-Clustering), which takes into account semantic concepts in the image together with the global image context for event representation.  

To the best of your knowledge, this is the first time that semantic concepts are used for image representation in egocentric videos and images.
With respect to our previous work published in \cite{talavera2015r-clustering}, we introduce the following contributions:
\begin{itemize}
	\item Methodology for egocentric photo streams description based on semantic information.
	\item Set of evaluation metrics applied to ground truth consistency estimation.
    \item Evaluation on an extended number of datasets, including our own, which will be published with this work.
    \item Exhaustive evaluation on a broader number of methods to compare with.
\end{itemize}


This manuscript is organized as follows: Section \ref{sec:approach} provides a  description of the proposed photo stream segmentation approach discussing the semantic and contextual features, the clustering and the graph-cut model. Section \ref{sec:results} presents  experimental results and, finally, Section \ref{sec:conclusions} summarizes the important outcomes of the proposed method providing some concluding remarks.


\section{SR-Clustering for Temporal Photo Stream Segmentation}
\label{sec:approach}
\begin{figure}[ht!]
\centering      
\includegraphics[width=\textwidth]{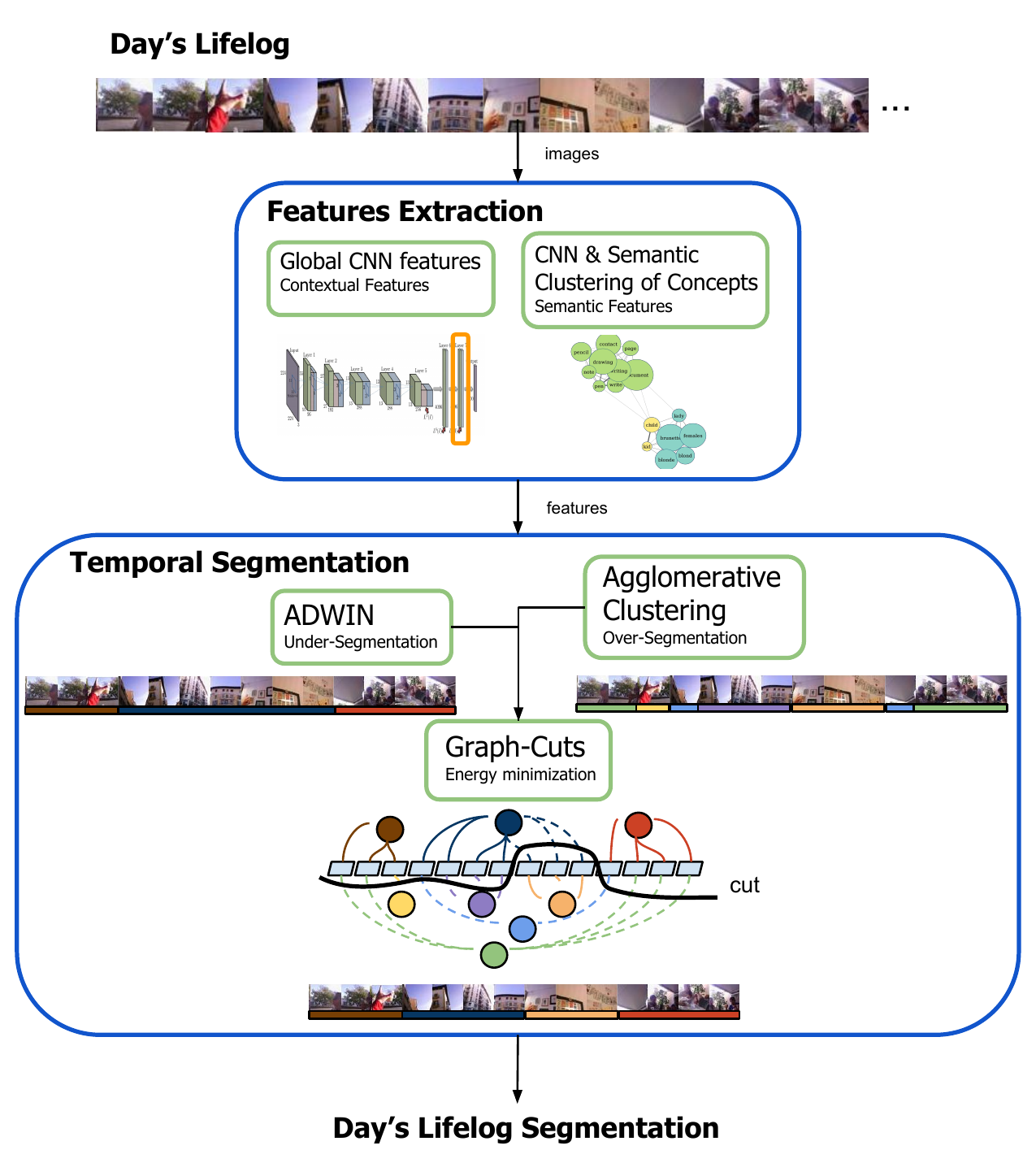}
\caption{General scheme of the Semantic Regularized Clustering (SR-Clustering) method.}
\label{fig:r_clustering_scheme}
\end{figure}

 A visual overview of the proposed method is given in Fig. \ref{fig:r_clustering_scheme}. The  input is a day long photo stream from which contextual and semantic features are extracted. An initial clustering is performed by AC and ADWIN. Later, GC is applied to look for a trade-off between the AC (represented by the bottom colored circles) and ADWIN (represented by the top colored circles) approaches. The binary term of the GC imposes smoothness and similarity of consecutive frames in terms of the CNN image features. The output of the proposed method is the segmented photo stream.  \textcolor{red}{In this section}, we introduce the semantic and contextual features of SR-clustering and provide a detailed description of the segmentation approach.

\subsection{Features}
  
We assume that two consecutive images belong to the same segment if they can be described by similar image features. When we refer to the features of an image, we usually consider low-level image features (e.g. color, texture, etc.) or a global representation of the environment (e.g. CNN features). However, the objects or concepts that semantically represent an event are also of high importance for the photo stream segmentation. Below, we detail the features that semantically describe the egocentric images. 

\subsubsection{Semantic Features}
 
Given an image $I$, let us consider a tagging algorithm that returns a set of objects/tags/concepts detected in the images with their associated confidence value. \textcolor{red}{The confidence values of each concept} form a semantic feature vector to be used for the photo streams segmentation. Usually, the number of concepts detected for each sequence of images is large (often, some dozens). Additionally, redundancies in the detected concepts are quite often due to the presence of synonyms or semantically related words. To manage the semantic redundancy, we will rely on WordNet \cite{miller1995wordnet}, which is a lexical database that groups English words into sets of synonyms, providing additionally short definitions and word relations.

\begin{figure}
\centering
\includegraphics[width=\textwidth]{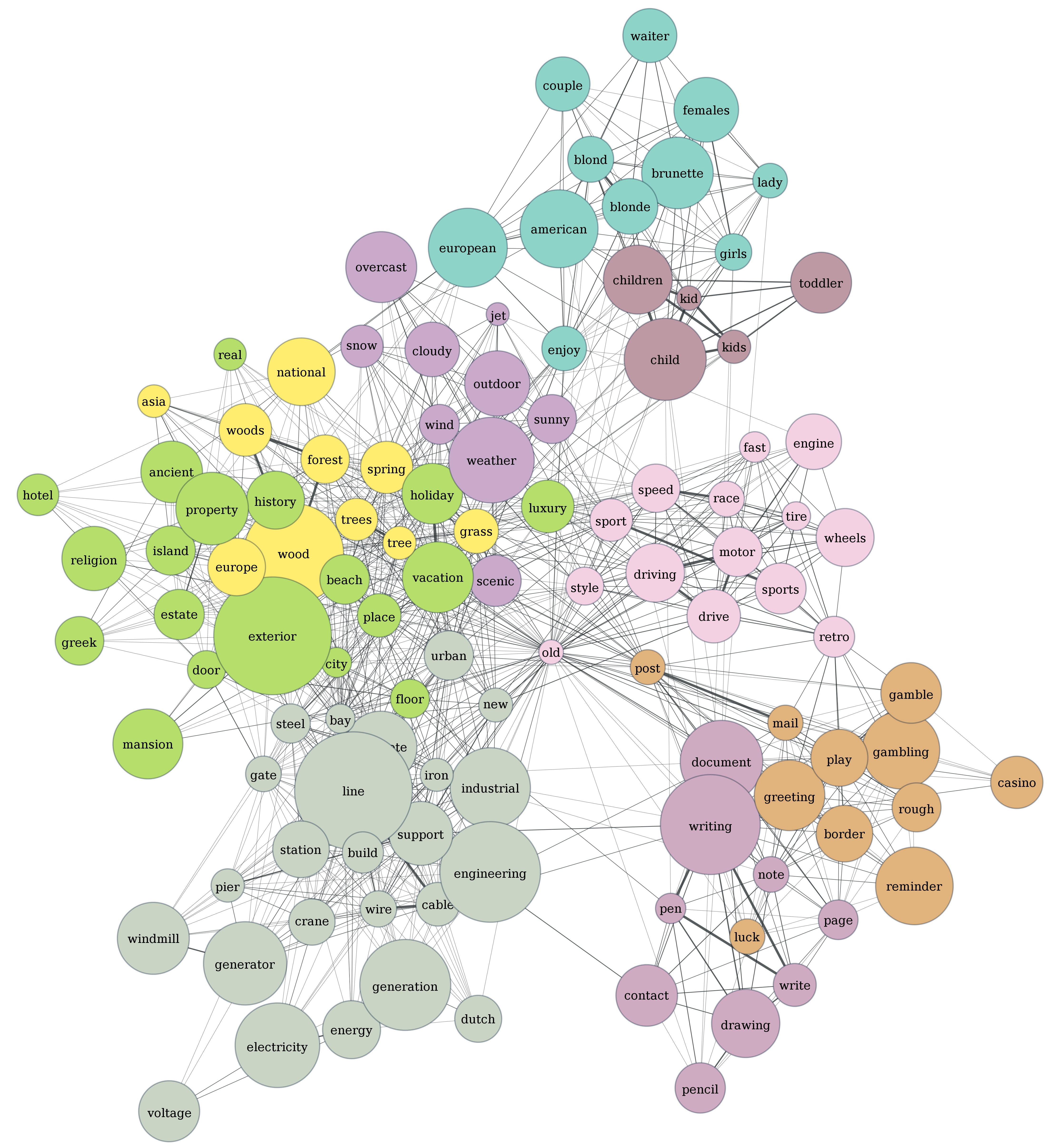}
\caption{Simplified graph obtained after calculating  similarities of the concepts of a day's lifelog and clustering them. Each color corresponds to a different cluster, the edge width represents the magnitude of the similarity between concepts, and the nodes size represents the number of connections they have (the biggest node in each cluster is the representative one). We only showed a small subset of the 100 clusters. This graph was drawn using graph-tool (\url{http://graph-tool.skewed.de}).}
\label{fig:tags_graph}
\end{figure}

Given a day's lifelog, let us cluster the concepts by relying on their synset ID in WordNet to compute their similarity in meaning, and following, apply clustering (e.g. Spectral clustering) to obtain 100 clusters. As a result, we can semantically describe each image in terms of 100 concepts and their associated confidence scores.
Formally, we first construct a semantic similarity graph $\mathcal{G} = \{V, E, W\}$, where each vertex or node $v_i \in V$ is a  concept, each edge $e_{ij} \in E$ represents a semantic relationship between two concepts,  $v_i$ and $v_j$ and each weight $w_{ij} \in W$ represents the strength of the semantic relationship, $e_{ij}$. We compute each $w_{ij}$ by relying on the meanings and the associated similarity given by WordNet,  between each appearing pair. To do so, we use the max-similarity between all the possible meanings $m_i^k$ and $m_j^r$ in $M_i$ and $M_j$ of the given pair of concepts $v_i$ and $v_j$:
\begin{equation*}
w_{ij} = \max_{m_i^k \in M_i, m_j^r \in M_j} sim(m_i^k, m_j^r).
\end{equation*}
To compute the Semantic Clustering, we use their similarity relationships in the spectral clustering algorithm to obtain 100 semantic concepts, $|C| = 100$. 
In Fig. \ref{fig:tags_graph}, a simplified example of the result obtained after the clustering procedure is shown. For instance, in the purple cluster, similar concepts like 'writing', 'document', 'drawing', 'write', etc. are grouped in the same cluster, and 'writing' is chosen as the most representative term. For each cluster, we choose as its representative concept, the one with the highest sum of similarities with the rest of elements in the cluster.

The semantic feature vector $f^s \in \mathbb{R}^{|C|}$ for image $I$ is a 100-dimensional array, such that each component $f^s(I)_j$ of the vector represents the confidence with which the j-th concept is detected in the image. The confidence value for the concept $j$, representing the cluster $C_j$, is obtained as the sum of the confidences $r_I$ of all the concepts included in $C_j$ that have also been detected on image $I$:
\begin{equation*}
	f^s(I)_j = \sum_{c_k \in \{C_j\}} r_I(c_k)
\end{equation*}
where $C_{I}$ is the set of concepts detected on image $I$, $C_j$ is the set of concepts in cluster $j$, and $r_I(c_k)$ is the confidence associated to concept $c_k$ on image $I$. The final confidence values are normalized so that they are in the interval $[0,1]$.

\begin{figure}[ht!]
        \centering
      \includegraphics[width=\textwidth]{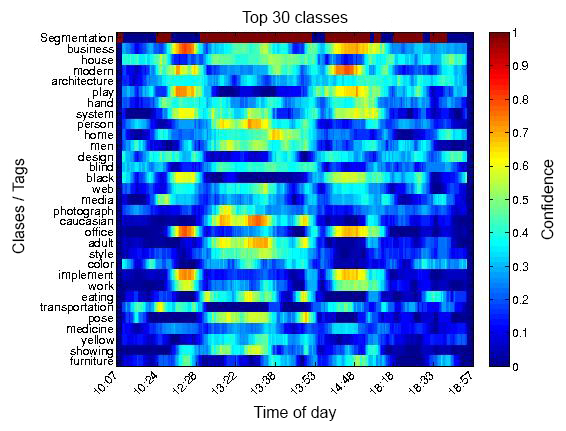}
        \caption{Example of the final semantic feature matrix obtained for an egocentric sequence. The top 30 concepts (rows) are shown for all the images in the sequence (columns). Additionally, the top row of the matrix shows the ground truth (GT) segmentation of the dataset.}
\label{fig:Estefania2_IMAGGA_Top30.jpg}
\end{figure}

Taking into account that the camera wearer can be continuously moving, even if in a single environment, the objects that can be appearing in temporally adjacent images may be different. To this end, we apply a Parzen Window Density Estimation method \cite{parzen1962estimation} to the matrix obtained by concatenating the semantic feature vectors along the sequence to obtain a smoothed and temporally coherent set of confidence values.
Additionally, 
we discard the concepts with a low variability of confidence values along the sequence which correspond to non-discriminative concepts that can appear on any environment. The low variability of confidence value of a concept may correspond to constantly having high or low confidence value in most environments.

\begin{figure}[ht!]
\centering
\includegraphics[width=1\textwidth]{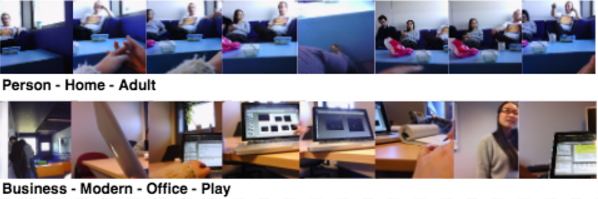}
\caption{Example of extracted tags on different segments. The first one corresponds to the period from 13.22 - 13.38 where the user is having lunch with colleagues, and the second, from 14.48 - 18.18, where he/she is working in the office with the laptop.}
\label{fig:tagspersceen}
\end{figure}

\begin{figure}[ht!]
\centering
\includegraphics[width=\textwidth]{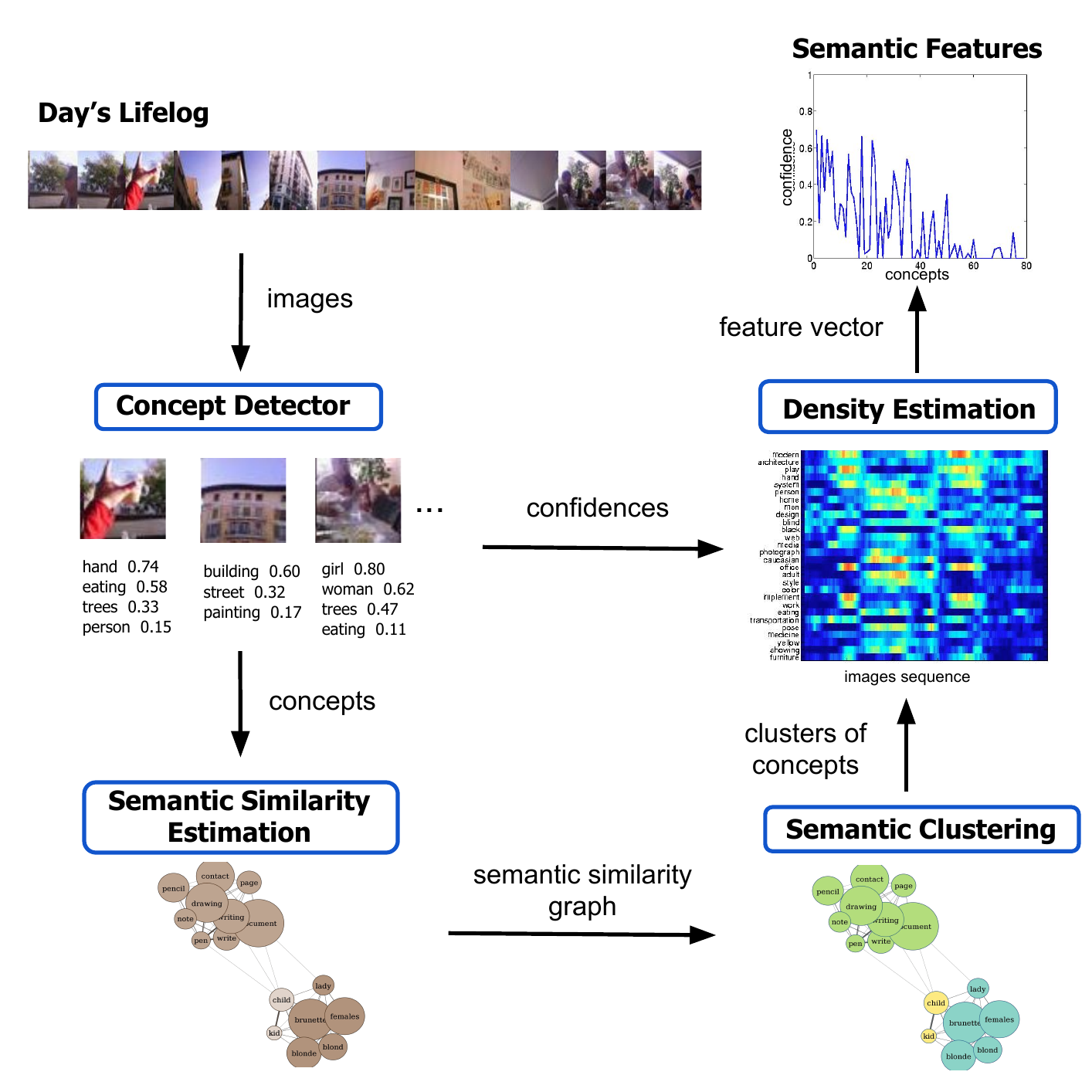}
\caption{General scheme of the semantic feature extraction methodology.}
\label{fig:semantic_features}
\end{figure}

In Fig. \ref{fig:Estefania2_IMAGGA_Top30.jpg}, the matrix of concepts (semantic features) associated to an egocentric sequence is shown, displaying only the top 30 classes. Each column of the matrix corresponds to a frame and each row indicates the confidence with which the concept is detected in each frame. In the first row, the ground truth of the temporal segmentation is shown for comparison purposes. With this representation, repeated patterns along a set of continuous images correspond to the set of concepts that characterizes an event. For instance, the first frames of the sequence represent an indoor scene, characterized by the presence of people (see examples Fig. \ref{fig:tagspersceen}). The whole process is summarized in Fig. \ref{fig:semantic_features}.

In order to consider the semantics of temporal segments, we used a concept detector based on the auto-tagging service developed by Imagga Technologies Ltd. Imagga's auto-tagging technology \footnote{\url{http://www.imagga.com/solutions/auto-tagging.html}} uses a combination of image recognition based on deep learning and CNNs using very large collections of human annotated photos. The advantage of Imagga's Auto Tagging API is that it can directly recognize over 2,700 different objects and in addition return more than 20,000  abstract concepts related to the analyzed images. 


\subsubsection{Contextual Features}  
In addition to the semantic features, we represent images with a feature vector extracted from a pre-trained CNN. The CNN model that we use for computing the images representation is the AlexNet, which is detailed in \cite{NIPS2012_4824}. The features are computed by removing the last layer corresponding to the classifier from the network. We used the deep learning framework Caffe \cite{Jia13caffe} in order to run the CNN. Due to the fact that the weights have been trained on the ImageNet database \cite{deng2009imagenet}, which is made of images containing single objects, we expect that the features extracted from images containing multiple objects will be representative of the environment. It is worth to remark that we did not use the weights obtained using a pre-trained CNN on the scenes from Places 205 database \cite{zhou2014learning}, since the Narrative camera's field of view  is narrow, which means that mostly its field-of-view is very restricted to characterize the whole scene. Instead, we usually only see objects on the foreground. As detailed in \cite{talavera2015r-clustering}, to reduce the large variation distribution of the CNN features, which results problematic when computing distances between vectors, we used a signed root normalization to produce more uniformly distributed data \cite{ZhengWHT14}.

\subsection{Temporal Segmentation}

The SR-clustering for temporal segmentation is based on fusing the semantic and contextual features with the R-Clustering method described in \cite{talavera2015r-clustering}.

\subsubsection{Agglomerative Clustering}
After the concatenation of semantic and contextual features, the hierarchical Agglomerative Clustering (AC) method is applied following a bottom-up clustering procedure. In each iteration, the method merges the most similar pair of clusters based on the distances among the image features, updating the elements similarity matrix. This is done until exhausting all possible consistent combinations. The {\em cutoff} global parameter defines the consistency of the merged clusters. We use the Cosine Similarity between samples, which is suited for high-dimensional positive spaces \cite{Tan:2005:IDM:1095618}. The shortcoming of this method is that it tends to over-segment the photo streams.

\subsubsection{ADWIN}
To compensate the over-segmentation produced by AC, we proposed to model the egocentric sequence as a multi-dimensional data stream and to detect changes in the mean distribution through an adaptive learning method called ADWIN \cite{bifet2007learning}, which provides a rigorous statistical guarantee of performance in terms of false positive rate. The method, based on the Hoeffding's inequality \cite{Hoeffding1963}, tests recursively if the difference between the averages of two temporally adjacent (sub)windows of the data, say $W_1$ and $W_2$, is larger than a threshold. The value of the threshold takes into account if both sub-windows are {\em large enough} and {\em distinct enough} for a $k-$dimensional signal  \cite{Drozdzal2014},  computed as:
\begin{equation*}
\epsilon_{cut} = k^{1/p}\sqrt{\frac{1}{2m}\ln\frac{4}{k\delta'}}
\end{equation*}
where $p$ indicates the $p-$norm, $\delta \in (0, 1)$ is a user defined confidence parameter, and $m$ is the harmonic mean between the lengths of  $W_1$ and $W_2$. In other words, given a predetermined confidence, ADWIN statistically guarantees that it  will find any major change in the data means.
Given a confidence value $\delta$, the higher the dimension $k$ is, the more samples $n$ the bound needs to reach assuming the same value of $\epsilon_{cut}$. The higher the norm used is,  the less important the dimensionality $k$ is. Since we model the sequence as a high dimensional data stream, ADWIN is unable to predict changes involving a relatively small number of samples, which often characterizes \textcolor{red}{Low Temporal Resolution (LTR)} egocentric data, leading to under-segmentation. Moreover, since it considers only the mean change, it is able to detect changes due to other statistics such as the variance.  

\subsubsection{Graph-Cuts regularization}
We use Graph-Cuts (GC) as a framework to integrate both of the previously described approaches, AC and ADWIN, to find a compromise between them that naturally leads to a temporally consistent result. GC is an energy-minimization technique that works by finding the minimum of an energy function usually composed of two terms: the {\em unary term} $U$, also called data term, that describes the relationship of the variables to a possible class and the {\em binary term} $V$, also called pairwise or regularization term, that describes the relationship between two neighboring samples (temporally close images) according to their feature similarity. The binary term smooths boundaries between similar frames, while the unary term keeps the cluster membership of each sequence frame according to its likelihood. In our problem, we defined the unary term as the sum of 2 parts ($U_{ac}(f_i)$ and $U_{adw}(f_i)$). Each of them expresses the likelihood of an image $I_i$ represented by the set of features $f_i$ to belong to segments coming from the corresponding previously applied segmentation methods. The energy function to be minimized is the following:

\begin{equation*}
    E(f) = \sum_{i}^{n} \Bigg[ (1-\omega_1) U_{ac}(f_i) + \omega_1 U_{adw}(f_i) \Bigg] + \omega_2 \sum_{i}^{n} \Bigg [ \frac{1}{|N_i|} \sum_{j\in N_i} V_{i,j}(f_i,f_j) \Bigg ]
\end{equation*}

where $f_i = \left [ f^c(I_i), f^s(I_i) \right ],i=\lbrace1,...,n \rbrace$ are the set of contextual $f^c$ and semantic image features $f^s$ for the $i$-th image, $N_i$ is a set of temporal neighbors centered at $i$, and $\omega_1$ and $\omega_2$ ($\omega_1, \omega_2 \in [0,1]$) are the unary and the binary weighting terms, respectively. We can improve the segmentation outcome of GC by defining how much weight do we give to the likelihood of each unary term and balancing the trade-off between the unary and the pairwise energies, respectively. The minimization is achieved through the max-cut algorithm, leading to a temporal  segmentation with similar frames having as large likelihood as possible to belong to the same segment, while maintaining segment boundaries in temporally neighboring images with high feature dissimilarity. 

More precisely, the unary energy is composed of two terms representing, each of them, the likelihoods of each sample to belong to each of the clusters (or decisions) obtained either applying ADWIN ($T_{adw}$) or AC ($T_{ac}$) respectively:
\begin{equation*}
	U_{ac}(f_i) = P_{ac}(f_i \in T_{ac}), \; \;
	U_{adw}(f_i) = P_{adw}(f_i \in T_{adw})
\end{equation*}
The pair-wise energy is defined as: 
\begin{equation*}
	V_{i,j}(f_i, f_n) = e^{-\text{dist}(f_i, f_j)}
\end{equation*}
An illustration of this process  is shown in Fig. \ref{fig:r_clustering_scheme}.


\section{Experiments and Validation}
\label{sec:results}


In this section, we discuss the datasets and the statistical evaluation measurements used to validate the proposed model and to compare it with the state-of-the-art methods. To sum up, we apply the following methodology for validation:

\begin{enumerate}
	\item Three different datasets acquired by 3 different wearable cameras are used for validation.
   \item The F-Measure is used as a statistical measure to compare the performance of different methods.
	\item Two consistency measures to compare different manual segmentations is applied.
	\item Comparison results of SR-Clustering with 3 state-of-the-art techniques is provided.
	\item Robustness of the final proposal is proven by validating the different components of SR-Clustering.
\end{enumerate}


\subsection{Data} 

To evaluate the performance of our method, we used 3 public datasets (EDUB-Seg, AIHS and Huji EgoSeg's sub dataset) acquired by three different wearable cameras (see Table \ref{datasets-table}).

\begin{table}[ht!]
\centering
\begin{tabular}{l|llllll}
\textbf{Dataset} & \multicolumn{1}{l}{\textbf{Camera}} & \multicolumn{1}{l}{\textbf{FR}} & \multicolumn{1}{l}{\textbf{SR}} & \multicolumn{1}{l}{\textbf{\#Us}} & \multicolumn{1}{l}{\textbf{\#Days}} & \textbf{\#Img} \\ \hline
EDUB & Narrative & 2 fpm & 2592x1944 & 7 & 20 & 18,735 \\
AIHS-subset & SenseCam & 3 fpm &  640x480 & 1 & 5 & 11,887 \\
Huji EgoSeg & GoPro Hero3+ & 30fps* & 1280x720 & 2 & 2 & 700
\end{tabular}
\caption{Table summarizing the main characteristics of the datasets used in this work: frame rate (FR), spatial resolution (SR), number of users (\#Us),  number of days (\#Days), number of images (\#Img). The Huji EgoSeg dataset has been subsampled to 2 fpm as detailed  in the main text.}
\label{datasets-table}
\end{table}

\textbf{EDUB-Seg}: \textcolor{red}{is a dataset acquired by people from our lab with the Narrative Clip, which takes a picture every 30 seconds. Our Narrative dataset, named EDUB-Seg (Egocentric Dataset of the University of Barcelona - Segmentation), contains a total of 18,735 images captured by 7 different users during overall 20 days. To ensure diversity, all users were wearing the camera in different contexts: while attending a conference, on holiday, during the weekend, and during the week. The EDUB-Seg dataset is an extension of the dataset used in our previous work \cite{talavera2015r-clustering}, that we call EDUB-Seg (Set1) to distinguish it from the newly added in this paper EDUB-Seg (Set2).}
The camera wearers, as well as all the researchers involved on this work, were required to sign an informed written consent containing set of moral principles \cite{wiles2008visual, kelly2013ethical}. Moreover, all researchers of the team have signed to do not publish any image identifying a person in a photo stream without his/her explicit permission, except unknown third parties. 

\textbf{AIHS \textcolor{red}{subset}}: is a subset of the daily images from the database called \emph{All I Have Seen} (AIHS) \cite{NIPS},  recorded by the SenseCam camera that takes a picture every 20 seconds.\textcolor{red}{The original AIHS dataset \footnote{http://research.microsoft.com/en-us/um/people/jojic/aihs/} has no timestamp metadata. We manually divided the dataset in five days guided by the pictures the authors show in the website of their project and based on the daylight changes observed in the photo streams. The five days sum up a total of 11,887 images.} 
 Comparing both cameras (Narrative and SenseCam), we can remark their difference with respect to the cameras' lens (fish eye vs normal), and the quality of the images they record. Moreover, SenseCam acquires images with a larger field of view and significant deformation and blurring. We manually defined the GT for this dataset following the same criteria we used for the EDUB-Seg photo streams.

\textbf{Huji EgoSeg}: due to the lack of other publicly available LTR datasets for event segmentation, we also test our temporal segmentation method to the ones provided in the dataset Huji EgoSeg \cite{polegtemporal}. This dataset was acquired by the GoPro camera, which captures videos with a temporal resolution of 30fps. Considering the very significant difference in frame rate of this camera compared to Narrative (2 fpm) and SenseCam (3 fpm), we applied a sub-sampling of the data by just keeping 2 images per minute, \textcolor{red}{to make} it comparable to the other datasets.  In this dataset, several short videos recorded by two different users are provided. Consequently, after sub-sampling all the videos, we merged the resulting images from all the short videos to construct a dataset per each user, which consists of a total number of 700 images. The images were merged following the numbering order that was provided by the authors to their videos.  We also manually defined the GT for this dataset following the same used criteria for the EDUB-Seg dataset.

In summary, we evaluate the algorithms on \textcolor{red}{27 days with a total of 31,322 images recorded by 10 different users}. All datasets contain a mixture of highly variable indoor and outdoor scenes with a large variety of objects. We make public the EDUB-Seg dataset\footnote{\url{http://www.ub.edu/cvub/dataset/}}, together with our GT segmentations of the datasets Huji EgoSeg and AIHS \textcolor{red}{subset}.  Additionally, we release the SR-Clustering ready-to-use complete code\footnote{\url{https://github.com/MarcBS/SR-Clustering}}. 

\subsection{Experimental setup}
Following \cite{li2013daily}, we measured the performances of our method by using the F-Measure (FM) defined as follows: 
\begin{equation*}
FM= 2\frac{RP}{R+P},
\end{equation*}
where $P$ is the precision defined as $(P=\frac{TP}{TP+FP})$ and $R$ is the recall, defined as $(R=\frac{TP}{TP+FN})$. $TP$, $FP$ and $FN$ are the number of true positives, false positives and false negatives of the detected segment boundaries of the photo stream. We define the FM, where we consider TPs the images that the model detects as boundaries of an event and that were close to the boundary image defined in the GT by the annotator (given a tolerance of 5 images \textcolor{red}{in both sides}). The FPs are the images detected as events delimiters, but that were not defined in the GT, and the FNs the lost boundaries  by the model that are indicated in the GT. Lower FM values represent a wrong boundary detection while higher values indicate a good segmentation. Having the ideal maximum value of 1, where the segmentation correlates completely with the one defined by the user.

\begin{figure}[ht!]
        \centering
      \includegraphics[width=\textwidth]{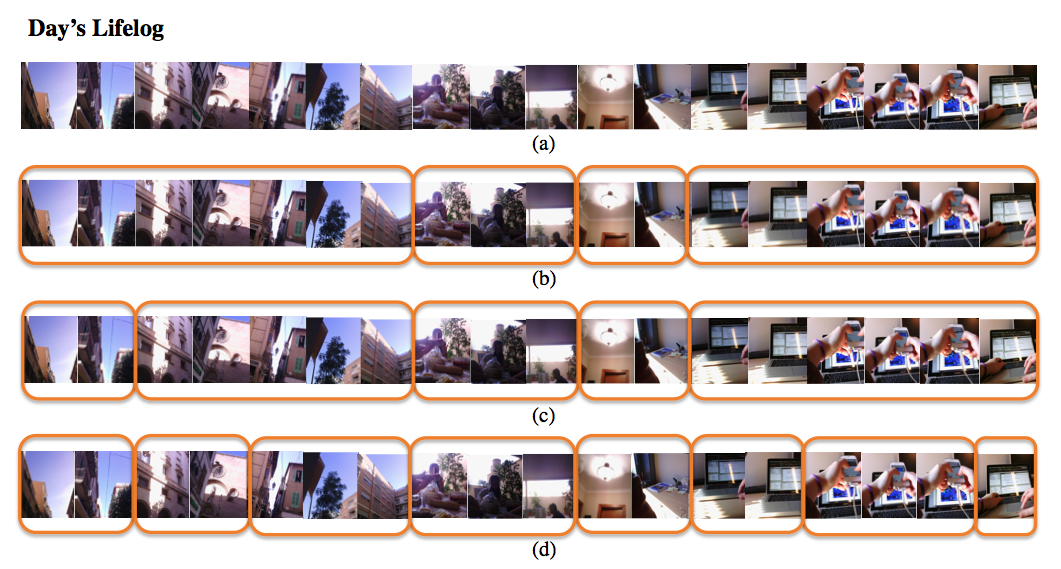}         \caption{Different segmentation results obtained by different subjects. (a) shows a part of a day. (b), (c) and (d) are examples of the segmentation performed by three different persons. (c) and (d) are refinements of the segmentation performed by (b). All three results can be considered as being correct, due to the subjective intrinsic of the  task. As a consequence, a segmentation consistency metric should not penalize different, yet consistent results of the segmentation.}
        \label{fig:imagenVariabilidad.jpg}
\end{figure}

The annotation of temporal segmentations of photo streams is a very subjective task. The fact that different users usually do not perform the same when annotating, may lead to bias in the evaluation performance. The problem of the subjectivity when defining the ground truth was previously addressed in the context of image segmentation \cite{Martin01adatabase}. In \cite{Martin01adatabase}, the authors proposed two measures to compare different segmentations of the same image. These measures are used to validate if the performed segmentations by different users are consistent and thus, can be served as an objective benchmark for the evaluation of the segmentation performances. In Fig. \ref{fig:imagenVariabilidad.jpg}, we report a visual example that illustrates the urge of employing this measure for temporal segmentation of egocentric photo streams. For instance,  the first segment in Fig. \ref{fig:imagenVariabilidad.jpg} is split in different segments when analyzed by different subjects although there is a degree of consistency among all segments. 
Inspired by this work, we re-define the local refinement error, between two temporal segments, as follows:
\begin{equation*}
E(S_A,S_B,I_i) = \frac{|R(S_A,I_i)\backslash R(S_B,I_i)|}{|R(S_A,I_i)|}, 
\label{eq:01}
\end{equation*}
\noindent where $\backslash$ denotes the set difference and, $S_A$ and $S_B$ are the two segmentations to be compared. $R(S_X,I_{i})$ is the set of images corresponding to the segment that contains the image $I_{i}$, when obtaining the segmentation boundaries $S_X$. 
 
If one temporal segment is a proper subset of the other, then the images lie in one interval of refinement, which results in the local error of zero. However, if there is no subset relationship, the two regions overlap in an inconsistent manner that results in a non-zero local error.
Based on the definition of local refinement we provided above, two error measures are defined by combining the values of the local refinement error for the entire sequence. The first error measure is called Global Consistency Error (GCE) that forces all local refinements to be in the same direction (segments of segmentation A can be only local refinements of segments of segmentation B). The second error measure is the Local Consistency Error (LCE), which allows refinements in different directions in different parts of the sequence (some segments of segmentation A can be of local refinements of segments of segmentation B and vice verse). The two measures are defined as follows: 
\begin{equation*}
GCE(S_A,S_B) = \frac{1}{n} min\{ \sum_{i}^n E(S_A,S_B,I_i),\sum_{i}^n E(S_B,S_A,I_i)\}
\label{eq:02}
\end{equation*}

\begin{equation*}
LCE(S_A,S_B) = \frac{1}{n} \sum_{i}^n min\{E(S_A,S_B,I_i),E(S_B,S_A,I_i)\}
\label{eq:03}
\end{equation*}

\noindent where $n$ is the number of images of the sequence, $S_A$ and $S_B$ are the two different temporal segmentations and $I_i$ indicates the i-th image of the sequence. The GCE and the LCE measures produce output values in the range $[0,1]$ where 0 means no error.

\begin{figure}[ht!]
        \centering
        \includegraphics[width=35em,height=13em]{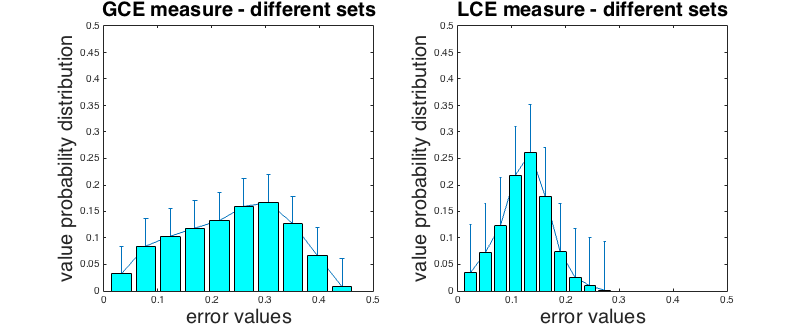}
		\includegraphics[width=35em,height=13em]{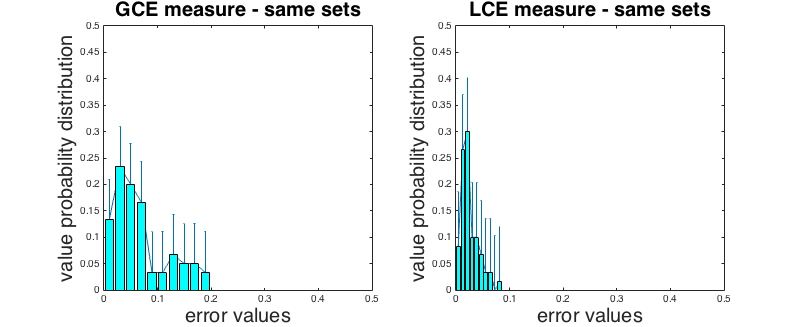}
        \caption{GCE (left) and LCE (right) normalized histograms with the error values distributions, showing  their mean and variance. The first row graphs represent the distribution of errors comparing segmentations of different sequences while the second row graphs show the distribution of error when comparing segmentations of the same set, including the segmentation of the camera wearer.}
\label{fig:ErrorMeasures_includingUser}
\end{figure}

\begin{figure}[ht!]
        \centering
        \includegraphics[width=35em,height=13em]{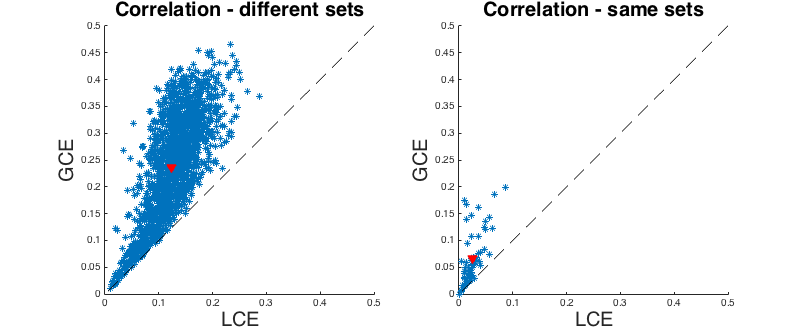}
        \caption{LCE vs GCE for pairs of segmentations of different sequences (left) and for pairs of segmentations of the same sequence (right). The differences w.r.t. the dashed line x=y show how GCE is a stricter measure than LCE. The red dot represents the mean of all the cloud of values, including the segmentation of the camera wearer.}
        \label{fig:ErrorMeasuresCorr_includingUser}
\end{figure}


\begin{figure}[ht!]
        \centering
        \includegraphics[width=35em,height=13em]{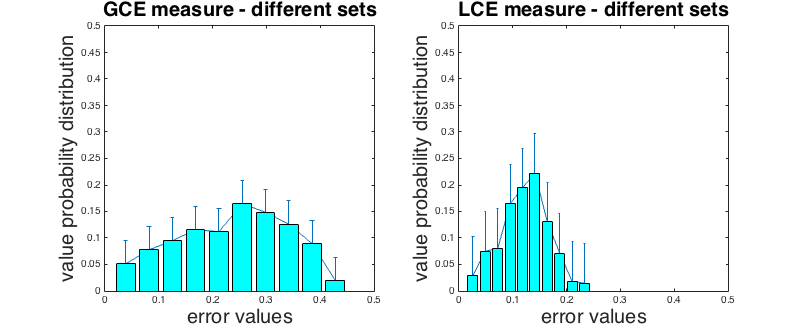}
		\includegraphics[width=35em,height=13em]{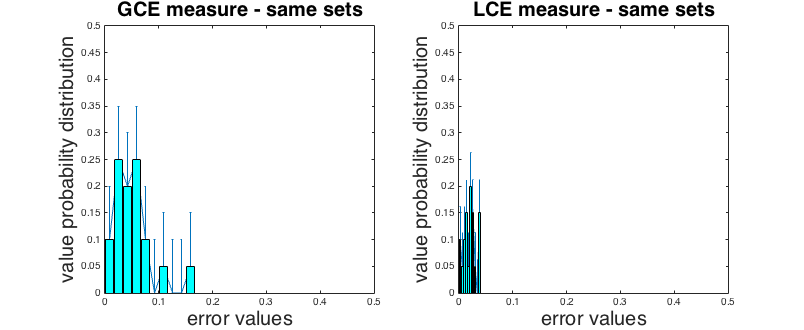}
        \caption{GCE (left) and LCE (right) normalized histograms with the error values distributions, showing their mean and variance. The first row graphs represent the distribution of the errors comparing segmentations of different sequences while the second row graphs show the distribution of the errors when comparing segmentations of the same set, excluding the segmentation of the camera wearer.}
\label{fig:ErrorMeasures_excludingUser}
\end{figure}

\begin{figure}[ht!]
        \centering
        \includegraphics[width=35em,height=13em]{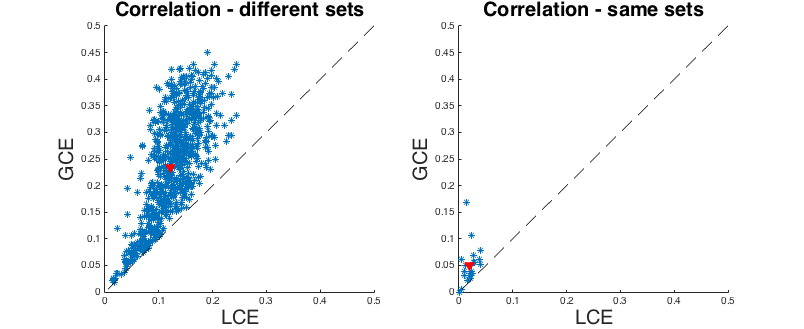}
        \caption{LCE vs GCE for pairs of segmentations of different sequences (left) and for pairs of segmentations of the same sequence (right). The differences w.r.t. the dashed line x=y show how GCE is a stricter measure than LCE. The red dot represents the mean of all the cloud of values, excluding the segmentation of the camera wearer.}
        \label{fig:ErrorMeasuresCorr_excludingUser}
\end{figure}


To verify that there is consistency among different people for the task of temporal segmentation, we asked three different subjects to segment each of the 20 sets of the EDUB-Seg dataset into events. The subjects were instructed to consider an \textit{event} as a semantically perceptual unit that can be inferred by visual features, without any prior knowledge of what the camera wearer is actually doing. No instructions were given to the subjects about the number of segments they should annotate. This process gave rise to 60 different segmentations. The number of all possible pairs of segmentations is 1800, 60 of which are pairs of segmentations of the same set. For each pair of segmentations, we computed GCE and LCE. First, we considered only pairs of segmentations of the same sequence and then, considered the rest of possible pairs of segmentations in the dataset. The first two graphics in Fig. \ref{fig:ErrorMeasures_includingUser} (first row) show the GCE (left) and LCE (right) when comparing each set segmentations with the segmentations applied on the rest of the sets. The two graphics in the second row show the distribution of the GCE (left) and LCE (right) error when analyzing different segments describing the same video. As expected, the distributions that compare the segmentations over the same photo stream have the center of mass to the left of the graph, which means that the mean error between the segmentations belonging to the same set is lower than the mean error between segmentations describing different sets. In Fig. \ref{fig:ErrorMeasuresCorr_includingUser} we compare, for each pair of segmentations, the measures produced by different datasets segmentations (left) and the measures produced by segmentations of the same dataset (right). In both cases, we plot LCE vs. GCE. 
\textcolor{red}{As expected,} the average error between segmentations of the same photo stream (right) is lower than the average error between segmentations of different photo streams (left). \textcolor{red}{Moreover, as indicated by the shape of the distributions on the second row of Fig.\ref{fig:ErrorMeasuresCorr_includingUser} (right), the peak of the LCE is very close to zero. } Therefore, we conclude that given the task of segmenting an egocentric photo stream into events, different people tend to produce consistent and valid segmentation. Fig. \ref{fig:ErrorMeasures_excludingUser} and \ref{fig:ErrorMeasuresCorr_excludingUser} show segmentation comparisons of three different persons (not being the camera wearer) that were asked to temporally segment a photo stream and confirm our statement that different people tend to produce consistent segmentations.

Since our interpretation of events is biased by our personal experience,  the segmentation done by the camera wearer could be very different by the segmentations done by third persons. To quantify this difference, in Fig. \ref{fig:ErrorMeasures_includingUser} and Fig. \ref{fig:ErrorMeasuresCorr_includingUser} we evaluated the LCE and the GCE including also the segmentation performed by the camera wearer. From this comparison, we can observe that the error mean does not vary but that the degree of local and global consistency is higher when the set of annotators does not include the camera wearer as it can be appreciated by the fact that the distributions are slightly shifted to the left and thinner. However, since this variation is of the order of $0.05\%$, we can conclude that event segmentation of egocentric photo streams can be objectively evaluated.


\begin{figure}[ht]
    \centering
    \includegraphics[width=1.0\linewidth]{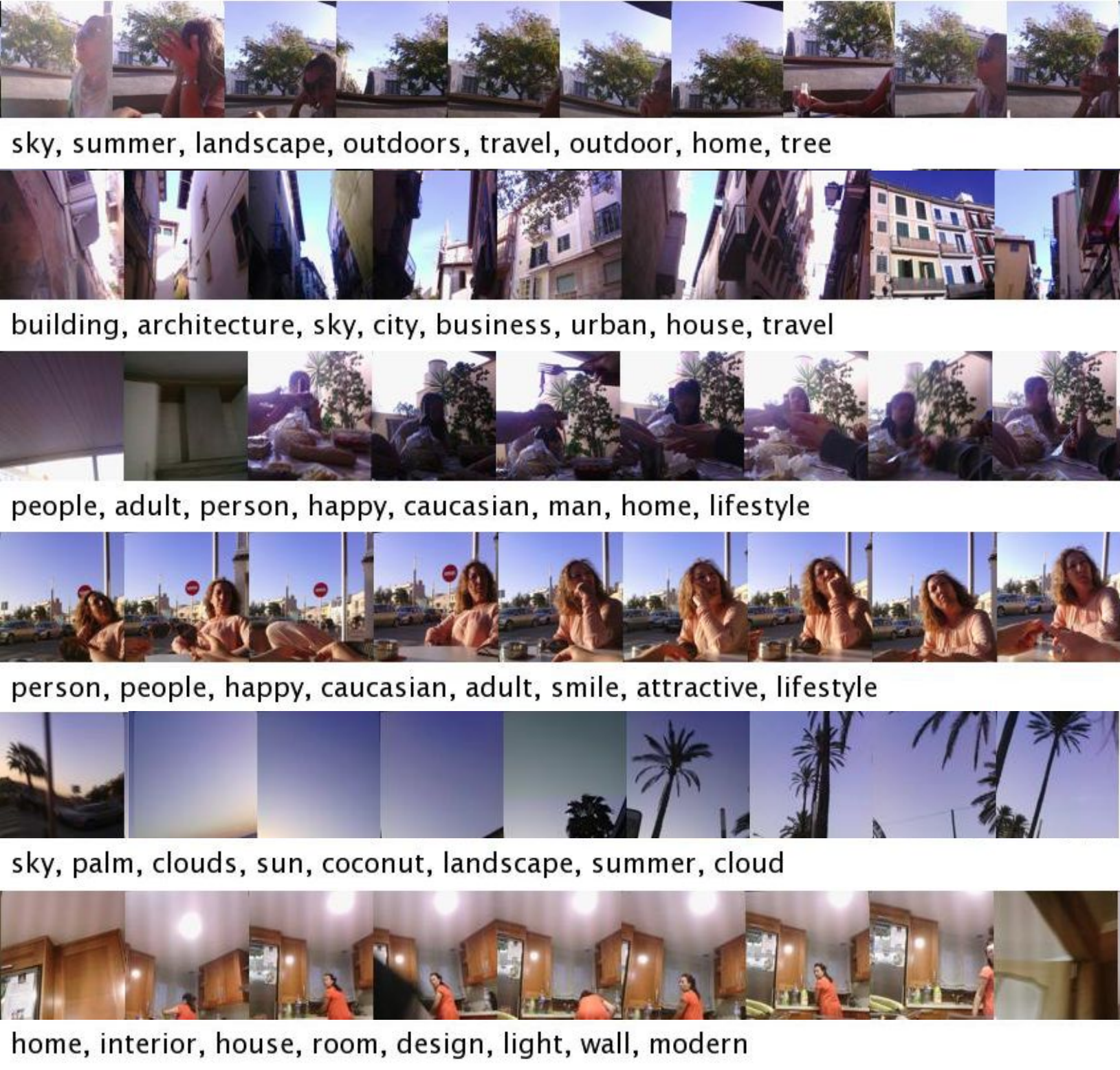}
    \caption{Illustration of our SR-Clustering segmentation results from a subset of pictures from a Narrative set. Each line represents a different segment. Below each segment we show the top 8 found concepts (from left to right). Only a few pictures from each segment are shown.}
    \label{fig:segmentation_results2}
\end{figure}

When comparing the different segmentation methods w.r.t. the obtained FM (see section \ref{sec:exp_results}), we applied a grid-search for choosing the best combination of hyper-parameters. The set of hyper-parameters tested are the following:
\begin{itemize}
\item AC linkage methods $\in$ \{\emph{ward, centroid, complete, weighted, single, median, average,}\}
\item AC cutoff $\in \{0.2,0.4,\dots,1.2\}$,
\item GraphCut unary weight $\omega_1$ and binary weight $\omega_2 \in \{0,0.1,0.2,\dots,1\}$,
\item AC-Color $t \in \{10,25,40,50,60,80,90,100\}$.
\end{itemize}


\subsection{Experimental results}\label{sec:exp_results}

In Table \ref{tab:results_summary}, we show the FM results obtained by different segmentation methods over different datasets. 
The first two columns correspond to the  datasets used  in \cite{talavera2015r-clustering}: AIHS-subset and  EDUB-Seg (Set1).  The third column corresponds to the EDUB-Seg (Set2) introduced in this paper. Finally, the fourth column corresponds to the results on the whole EDUB-Seg.
The first part of the table (first three rows) presents comparisons to  state-of-the-art methods. The second part of the table (next 4 rows),  shows comparisons to  different components of our proposed clustering method with and without semantic features. Finally, the third part of the table shows the results obtained using different variations of our method.
\begin{table}[!h] 
\centering
\small
\bgroup
\def\arraystretch{1.25}
\begin{tabular}{lrrrr}
  & \hspace*{-3mm} AIHS \cite{NIPS} & EDUB-Seg Set1 & EDUB-Seg Set2 & EDUB-Seg \\ \hline 
 Motion \cite{bolanos2014video} & 0.66 & 0.34 & & \\
 AC-Color \cite{grauman2015} & 0.60 & 0.37 & \textcolor{red}{0.54} & \textcolor{red}{0.50} \\ 
 R-Clustering \cite{talavera2015r-clustering} & \textbf{0.79} & 0.55 & & \\ \hdashline
 ADW & 0.31 & 0.32 & & \\
 ADW-ImaggaD & 0.35 & 0.55 & \textcolor{red}{0.29} & \textcolor{red}{0.36} \\
 AC & 0.68 & 0.45 & & \\
 AC-ImaggaD & 0.72 & 0.53 & \textcolor{red}{0.64} & \textcolor{red}{0.61} \\ \hdashline
 SR-Clustering-LSDA & 0.78 & 0.60 & \textcolor{red}{0.64} & \textcolor{red}{0.61} \\
 SR-Clustering-NoD & 0.77 & 0.66 & \textcolor{red}{0.63} & \textcolor{red}{0.60}  \\  
\textbf{SR-Clustering} & 0.78 & \textbf{0.69} & \textbf{\textcolor{red}{0.69}} & \textbf{\textcolor{red}{0.66}}\\
\end{tabular}
\egroup
\caption{Average FM results of the state-of-the-art works on the egocentric datasets (first part of the table); for each of the components of our method (second part); and for each of the variations of our method (third part). The last line shows the results of our complete method. AC stands for Agglomerative Clustering, ADW for ADWIN and ImaggaD is our proposal for semantic features, where D stands for Density Estimation.}
\label{tab:results_summary}
\end{table}

In the first part of Table \ref{tab:results_summary}, we compare to state-of-the-art methods. The first method is the Motion-Based segmentation algorithm proposed by Bola\~nos et al. \cite{bolanos2014video}. As can be seen, the average results obtained are far below SR-Clustering. This can be explained by the type of features used by the method, which are more suited for applying a motion-based segmentation. This kind of segmentation is more oriented to recognize activities and thus, is not always fully aligned with the event segmentation labeling we consider (i.e. in an event where the user goes outside of a building, and then enters to the underground tunnels can be considered "in transit" by the Motion-Based segmentation, but be considered as three different events in our event segmentation). Furthermore, the obtained FM score on the Narrative datasets is lower than the SenseCam's for several reasons: Narrative has lower frame rate compared to Sensecam (AIHS dataset), which is a handicap when computing motion information, and a narrower field of view, which decreases the semantic information present in the image. We also evaluated the proposal of Lee and Grauman \cite{grauman2015} (best with $t=25$), where they apply an Agglomerative Clustering segmentation using LAB color histograms. In this case, we see that the algorithm is even far below the obtained results by AC, where the Agglomerative Clustering algorithm is used over contextual CNN features instead of colour histograms. The main reason for this performance difference comes from the high difference in features expressiveness, that supports the necessity of using a rich set of features for correctly segmenting highly variable egocentric data.
The last row of the first section of the table shows the results obtained by our previously published method \cite{talavera2015r-clustering}, where we were able to outperform the state-of-the-art of egocentric segmentation using contextual CNN features both on AIHS-subset and on EDUB-Seg Set1. Another possible method to compare with would be the one from Castro et al. \cite{CastroHickson2015}, although the authors do not provide their trained model for applying this comparison.

In the second part of Table \ref{tab:results_summary}, we compare the results obtained using only ADWIN or only AC with (ADW-ImaggaD, AC-ImaggaD) and without (ADW, AC) semantic features. One can see that the proposed semantic features, leads to an improved performance, indicating that these features are rich enough to provide improvements on egocentric photo stream segmentation.

Finally, on the third part of Table \ref{tab:results_summary}, we compared our segmentation methodology  using different definitions for the semantic features. In the SR-Clustering-LSDA case, we used a simpler semantic features description, formed by using the weakly supervised concept extraction method proposed in \cite{hoffman2014lsda}, namely LSDA. In the last two lines, we tested the model using our proposed semantic methodology (Imagga's tags) either without Density Estimation, SR-Clustering-NoD or with the final Density Estimation (SR-Clustering), respectively.

Comparing the results of SR-Clustering and R-Clustering on the first two datasets (AIHS-subset and EDUB-Seg Set1), we can see that our new method is able to outperform the results adding 14 points of improvement to the FM score, while keeping nearly the same FM value on the SenseCam dataset. The improvement achieved using semantic information can be also corroborated, when comparing the FM scores obtained on the second half of EDUB-Seg dataset (Set2 on the 3rd column) and on the complete version of this data (see the last column of the Table).

\begin{table}[!h] 
\centering
\bgroup
\def\arraystretch{1}
\begin{tabular}{lrr}
  & Huji EgoSeg \cite{polegtemporal} LTR  \\ \hline 
 ADW-ImaggaD & 0.59\\
 AC-ImaggaD & \textbf{0.88}\\ \hdashline
 \textbf{SR-Clustering} & \textbf{0.88}\\
\end{tabular}
\egroup
\caption{Average FM score on each of the tested methods using our proposal of semantic features on the dataset presented in \cite{polegtemporal}.}
\label{tab:results_summary_peleg}
\end{table}

In Table \ref{tab:results_summary_peleg} we report the FM score obtained by applying our proposed method on the sub-sampled Huji EgoSeg dataset to be comparable to LTR cameras. Our proposed method achieves a high performance, being 0.88 of FM for both AC and SR-Clustering when using the proposed semantic features. The improvement of the results when using the GoPro camera with respect to Narrative or SenseCam can be explained by two key factors: 1) the difference in the field of view captured by GoPro (up to 170{$^{\circ}$}) compared to SenseCam (135{$^{\circ}$}) and Narrative (70{$^{\circ}$}), 2) the better image quality achieved by the head mounted camera.

In addition to the FM score, we could not consider the GCE and LCE measures to compare the consistency of the automatic segmentations to the ground truth, since both methods lead to a number of segments much larger than the number of segments in the ground truth and therefore these measures would not descriptive enough.
This is due to the fact that any segmentation is a refinement of one segment for the entire sequence, and one image per segment is a refinement of any segmentation. Consequently, these two trivial segmentations, one segment for the entire sequence and one image per segment, achieve error zero for LCE and GCE. 
However, we observed that on average, the number of segments obtained by the method of Lee and Grauman \cite{grauman2015} is about 4 times bigger than the number of segments we obtained for the SenseCam dataset and about 2 times bigger than for the Narrative datasets. 
Indeed, we achieve an higher FM score with respect to the method of Lee and Grauman \cite{grauman2015}, since it produces a considerable over-segmentation. 


\subsection{Discussion}

The experimental results detailed in section \ref{sec:exp_results} have shown the advantages of using semantic features for the temporal segmentation of egocentric photo streams.
Despite the common agreement about the inability of low-level features in providing understanding of the semantic structure present in complex  events \cite{habibian2014recommendations}, and the need of semantic indexing and  browsing systems, the use of high level features in the context of egocentric temporal segmentation and summarization has been very limited.
This is mainly due to the difficulty of dealing with the huge variability of object appearance and illumination conditions in egocentric images. In the works of Doherty et al. \cite{Doherty:2008} and Lee and Grauman \cite{grauman2015}, 
 temporal segmentation is still based on  low level features. In addition to the difficulty of reliably recognizing objects, the temporal segmentation of egocentric photo streams has to cope with the lack of temporal coherence, which in practice means that motion features cannot reliably be estimated. The work of Castro et al. \cite{CastroHickson2015} relies on the visual appearance of single images to predict the activity class of an image and on meta-data such as the day of the week and hour of the day to regularize over time. However, due to the huge variability in appearance and timing of daily activities, this approach cannot be easily generalized to different users, implying that for each new user re-training of the model and thus, labeling of thousand of images is required.

The method proposed in this paper offers the advantage of being needless of a cumbersome learning stage and offers a better generalization. The employed concept detector, has been proved to offer a rich vocabulary to describe the environment surrounding the user. This rich characterization is not only useful for better segmentation of sequences into meaningful and distinguishable events, but also serves as a basis for event classification or activity recognition among others. 
For example,  Aghaei et al. \cite{BOT,aghaei2015towards,aghaei2016whom} employed the temporal segmentation method in \cite{talavera2015r-clustering} to extract and select segments with trackable people to be processed. However, incorporating the semantic temporal segmentation proposed in this paper, would allow, for example,  to classify  events into social or non-social events. Moreover, using additional existing semantic features in a scene may be used to differentiate between different types of a social event ranging from a official meeting (including semantics such as laptop, paper, pen, etc.) to a friendly coffee break (coffee cup, cookies, etc.).
Moreover, the semantic temporal segmentation proposed in this paper is useful for indexing and browsing.



\section{Conclusions and future work}
\label{sec:conclusions}

This paper proposed an unsupervised approach for the temporal segmentation of egocentric photo streams that is able to partition a day's lifelog in segments sharing semantic attributes, hence providing a basis for semantic indexing and event recognition. The proposed approach first detects concepts for each image separately by employing a CNN approach and later, clusters the detected concepts in a semantic space, hence defining the vocabulary of concepts of a day. Semantic features are combined with global image features capturing more generic contextual information to increase their discriminative power. By relying on these semantic features, a GC technique is used to integrate a statistical bound produced by the concept drift method, ADWIN and the AC, two methods with complementary properties for temporal segmentation. We evaluated the performance of the proposed approach on different segmentation techniques and on 17 day sets acquired by three different wearable devices, and we showed the improvement of the proposed method with respect to the state-of-the-art. Additionally, we introduced two consistency measures to validate the consistency of the ground truth. Furthermore, we made publicly available our dataset EDUB-Seg, together with the ground truth annotation and the code. We  demonstrated that the use of semantic information on egocentric data is crucial for the development of a high performance method.

Further research will be devoted to exploit the semantic information that characterizes the segments for event recognition, where social events are of special interest. Additionally, we are interested in using semantic attributes to describe the camera wearer context. Hence, opening new opportunities for development of systems that can take benefit from contextual awareness, including systems for stress monitoring and daily routine analysis.

\section*{Acknowledgments} 
\vspace*{-0.25cm} This work was partially founded by TIN2012-38187-C03-01, SGR 1219 and  grant to research project 20141510 to Maite Garolera (from Fundaci\'{o} Marat\'{o} TV3). The funders had no role in the study design, data collection and analysis, decision to publish, or preparation of the manuscript. M. Dimiccoli is supported by a \textit{Beatriu de Pin\'{o}s} grant (Marie-Curie COFUND action). P. Radeva is partly supported by an \textit{ICREA Academia'2014} grant.

\section*{References} 
\bibliographystyle{plain}
{\footnotesize
\bibliography{Bibliography}}

\begin{thebibliography}{10}

\bibitem{aghaei2015towards}
M.~Aghaei, M.~Dimiccoli, and P.~Radeva.
\newblock Towards social interaction detection in egocentric photo-streams.
\newblock In {\em Eighth International Conference on Machine Vision}, pages
  987514--987514. International Society for Optics and Photonics, 2015.

\bibitem{BOT}
M.~Aghaei, M.~Dimiccoli, and P.~Radeva.
\newblock Multi-face tracking by extended bag-of-tracklets in egocentric
  videos.
\newblock {\em Computer Vision and Image Understanding, Special Issue on
  Assistive Computer Vision and Robotics}, 149:146--156, 2016.

\bibitem{aghaei2016whom}
M.~Aghaei, M.~Dimiccoli, and P.~Radeva.
\newblock With whom do {I} interact? detecting social interactions in
  egocentric photo-streams.
\newblock In {\em Proceedings of the International Conference on Pattern
  Recognition}, 2016.

\bibitem{bifet2007learning}
A.~Bifet and R.~Gavalda.
\newblock Learning from time-changing data with adaptive windowing.
\newblock In {\em Proceedings of SIAM International Conference on Data Mining},
  2007.

\bibitem{bolanos2015towards}
M.~Bola{\~n}os, M.~Dimiccoli, and P.~Radeva.
\newblock Towards storytelling from visual lifelogging: An overview.
\newblock {\em To appear on IEEE Transactions on Human-Machine Systems}, 2016.

\bibitem{bolanos2014video}
M.~Bola{\~n}os, M.~Garolera, and P.~Radeva.
\newblock Video segmentation of life-logging videos.
\newblock In {\em Articulated Motion and Deformable Objects}, pages 1--9.
  Springer-Verlag, 2014.

\bibitem{bolanos2015visual}
M.~Bola{\~n}os, E.~Talavera R.~Mestre, X.~Gir{\'o} i~Nieto, and P.~Radeva.
\newblock Visual summary of egocentric photostreams by representative
  keyframes.
\newblock {\em arXiv preprint arXiv:1505.01130}, 2015.

\bibitem{boykov2001fast}
Y.~Boykov, O.~Veksler, and R.~Zabih.
\newblock Fast approximate energy minimization via graph cuts.
\newblock {\em IEEE Transactions on Pattern Analysis and Machine Intelligence},
  23(11):1222--1239, 2001.

\bibitem{CastroHickson2015}
D.~Castro, S.~Hickson, V.~Bettadapura, E.~Thomaz, G.~Abowd, H.~Christensen, and
  I.~Essa.
\newblock Predicting daily activities from egocentric images using deep
  learning.
\newblock In {\em proceedings of the 2015 ACM International symposium on
  Wearable Computers}, pages 75--82. ACM, 2015.

\bibitem{deng2009imagenet}
J.~Deng, W.~Dong, R.~Socher, L.-J. Li, K.~Li, and L.~Fei-Fei.
\newblock Imagenet: A large-scale hierarchical image database.
\newblock In {\em Proceedings of IEEE Conference on Computer Vision and Pattern
  Recognition}, pages 248--255, 2009.

\bibitem{Doherty:2008}
A.~R. Doherty and A.~F. Smeaton.
\newblock Automatically segmenting lifelog data into events.
\newblock In {\em Proceedings of the 2008 Ninth International Workshop on Image
  Analysis for Multimedia Interactive Services}, pages 20--23, 2008.

\bibitem{AJM-Wearablecameras}
Aiden~R. Doherty, Steve~E. Hodges, Abby~C. King, and et~al.
\newblock Wearable cameras in health: the state of the art and future
  possibilities.
\newblock In {\em American journal of preventive medicine}, volume~44, pages
  320--323. Springer, 2013.

\bibitem{Drozdzal2014}
M.~Drozdzal, J.~Vitria, S.~Segui, C.~Malagelada, F.~Azpiroz, and P.~Radeva.
\newblock Intestinal event segmentation for endoluminal video analysis.
\newblock In {\em Proceedings of International Conference on Image Processing},
  2014.

\bibitem{habibian2014recommendations}
A.~Habibian and C.~Snoek.
\newblock Recommendations for recognizing video events by concept vocabularies.
\newblock {\em Computer Vision and Image Understanding}, 124:110--122, 2014.

\bibitem{Hoeffding1963}
W.~Hoeffding.
\newblock Probability inequalities for sums of bounded random variables.
\newblock {\em Journal of the American Statistical Association}, 58(301):pp.
  13--30, 1963.

\bibitem{hoffman2014lsda}
J.~Hoffman, S.~Sergio, E.~S. Tzeng, R.~Hu, R.~Girshick J.~Donahue, T.~Darrell,
  and K.~Saenko.
\newblock Lsda: Large scale detection through adaptation.
\newblock In {\em Advances in Neural Information Processing Systems}, pages
  3536--3544, 2014.

\bibitem{Jia13caffe}
Y.~Jia.
\newblock {Caffe}: An open source convolutional architecture for fast feature
  embedding.
\newblock http://caffe.berkeleyvision.org/, 2013.

\bibitem{NIPS}
N.~Jojic, A.~Perina, and V.~Murino.
\newblock Structural epitome: a way to summarize one's visual experience.
\newblock pages 1027--1035, 2010.

\bibitem{kelly2013ethical}
P.~Kelly, S.~Marshall, H.~Badland, J.~Kerr, M.~Oliver, A.~Doherty, and
  C.~Foster.
\newblock An ethical framework for automated, wearable cameras in health
  behavior research.
\newblock {\em American journal of preventive medicine}, 44(3):314--319, 2013.

\bibitem{NIPS2012_4824}
A.~Krizhevsky, I.~Sutskever, and G.~E. Hinton.
\newblock Imagenet classification with deep convolutional neural networks.
\newblock In {\em Advances in Neural Information Processing Systems}, pages
  1097--1105. 2012.

\bibitem{grauman2015}
YJ. Lee and K.~Grauman.
\newblock Predicting important objects for egocentric video summarization.
\newblock {\em International Journal of Computer Vision}, 114(1):38--55, 2015.

\bibitem{li2013daily}
Z.~Li, Z.~Wei, W.~Jia, and M.~Sun.
\newblock Daily life event segmentation for lifestyle evaluation based on
  multi-sensor data recorded by a wearable device.
\newblock In {\em Proceedings of Engineering in Medicine and Biology Society},
  pages 2858--2861. IEEE, 2013.

\bibitem{kmean}
W.-H. Lin and A.~Hauptmann.
\newblock Structuring continuous video recording of everyday life using
  time-constrained clustering.
\newblock {\em Proceedings of SPIE, Multimedia Content Analysis, Management,
  and Retrieval}, 959, 2006.

\bibitem{egovideo}
Z.~Lu and K.~Grauman.
\newblock Story-driven summarization for egocentric video.
\newblock In {\em Proceedings of IEEE Conference on Computer Vision and Pattern
  Recognition}, pages 2714--2721, 2013.

\bibitem{Martin01adatabase}
D.~Martin, C.~Fowlkes, D.~Tal, and J.~Malik.
\newblock A database of human segmented natural images and its application to
  evaluating segmentation algorithms and measuring ecological statistics.
\newblock In {\em Proceedings of 8th International Conference on Computer
  Vision}, pages 416--423, 2001.

\bibitem{miller1995wordnet}
G.~A. Miller.
\newblock Wordnet: a lexical database for english.
\newblock {\em Communications of the ACM}, 38(11):39--41, 1995.

\bibitem{parzen1962estimation}
E.~Parzen.
\newblock On estimation of a probability density function and mode.
\newblock {\em The annals of mathematical statistics}, pages 1065--1076, 1962.

\bibitem{polegtemporal}
Y.~Poleg, C.~Arora, and S.~Peleg.
\newblock Temporal segmentation of egocentric videos.
\newblock In {\em Proceedings of the IEEE Conference on Computer Vision and
  Pattern Recognition}, pages 2537--2544, 2014.

\bibitem{PolegEP015}
Y.~Poleg, A.~Ephrat, S.~Peleg, and C.~Arora.
\newblock Compact {CNN} for indexing egocentric videos.
\newblock {\em CoRR}, abs/1504.07469, 2015.

\bibitem{talavera2015r-clustering}
E.~Talavera, M.~Dimiccoli, M.~Bolanos, M.~Aghaei, and P.~Radeva.
\newblock R-clustering for egocentric video segmentation.
\newblock In {\em Iberian Conference on Pattern Recognition and Image
  Analysis}, pages 327--336. Springer, 2015.

\bibitem{Tan:2005:IDM:1095618}
P.~N. Tan, M.~Steinbach, and V.~Kumar.
\newblock {\em Introduction to Data Mining, (First Edition)}.
\newblock Addison-Wesley Longman Publishing Co., Inc., Boston, MA, USA, 2005.

\bibitem{wiles2008visual}
R.~Wiles, J.~Prosser, A.~Bagnoli, A.~Clark, K.~Davies, S.~Holland, and
  E.~Renold.
\newblock Visual ethics: Ethical issues in visual research.
\newblock 2008.

\bibitem{ZhengWHT14}
L.~Zheng, Sh. Wang, F.~He, and Q.~Tian.
\newblock Seeing the big picture: Deep embedding with contextual evidences.
\newblock {\em CoRR}, abs/1406.0132, 2014.

\bibitem{zhou2014learning}
B.~Zhou, A.~Lapedriza, J.~Xiao, A.~Torralba, and A.~Oliva.
\newblock Learning deep features for scene recognition using places database.
\newblock In {\em Advances in Neural Information Processing Systems}, pages
  487--495, 2014.

\end{thebibliography}

\end{document}